\documentclass[journal]{IEEEtran}

\usepackage{mathtools}
\usepackage{nccmath}
\usepackage{color}
\usepackage{array}
\usepackage{stfloats}
\usepackage{nccmath}
\usepackage{setspace}
\usepackage{mwe}
\usepackage{microtype}
\usepackage[linesnumbered,ruled]{algorithm2e}
\usepackage{algorithmicx}
\usepackage{amsmath,amssymb,amsfonts}
\usepackage{xspace}
\usepackage{textcomp}
\usepackage{xcolor}
\usepackage{graphicx}
\usepackage{booktabs}
\usepackage{subcaption}
\newcommand{\argmin}{\mathop{\mathrm{argmin}}\limits} 
\newcommand{\argmax}{\mathop{\mathrm{argmax}}\limits}

\renewcommand\baselinestretch{1.00}

\begin{document}
\title{Parallelized and Randomized Adversarial Imitation Learning for Safety-Critical Self-Driving Vehicles}

\author{Won Joon Yun, MyungJae Shin, Soyi Jung, Sean Kwon, and Joongheon Kim
	\thanks{Preliminary versions were appeared in Proceedings of the International Joint Conference on Artificial Intelligence (IJCAI)~\cite{shin2019rail} and presented at \textit{ICML Workshop on AI for Autonomous Driving 2019}~\cite{shin2019icml}.}
	\thanks{W.J. Yun and J. Kim are with the Department of Electrical and Computer Engineering, Korea University, Seoul, Korea e-mails: ywjoon95@korea.ac.kr, joongheon@korea.ac.kr.}
	\thanks{M. Shin is with Mofl Inc., Daejeon, Korea e-mail: mjshin.cau@gmail.com.}
	\thanks{S. Jung is with the School of Software, Hallym University, Chuncheon, Korea e-mail: sjung@hallym.ac.kr.}
	\thanks{S. Kwon is with the Department of Electrical Engineering, California State University, Long Beach, CA, USA e-mail: sean.kwon@csulb.edu.}
	\thanks{W.J. Yun and M. Shin are equally contributed to this work (first authors).}
	\thanks{S. Jung and J. Kim are the corresponding authors of this paper.}
}

\maketitle

\begin{abstract}
Self-driving cars and autonomous driving research has been receiving considerable attention as major promising prospects in modern artificial intelligence applications. According to the evolution of advanced driver assistance system (ADAS), the design of self-driving vehicle and autonomous driving systems becomes complicated and safety-critical. In general, the intelligent system simultaneously and efficiently activates ADAS functions. Therefore, it is essential to consider reliable ADAS function coordination to control the driving system, safely.
In order to deal with this issue, this paper proposes a randomized adversarial imitation learning (RAIL) algorithm. The RAIL is a novel derivative-free imitation learning method for autonomous driving with various ADAS functions coordination; and thus it imitates the operation of decision maker that controls autonomous driving with various ADAS functions.
The proposed method is able to train the decision maker that deals with the LIDAR data and controls the autonomous driving in multi-lane complex highway environments. The simulation-based evaluation verifies that the proposed method achieves desired performance.
\end{abstract}


\IEEEpeerreviewmaketitle

\section{Introduction}
\IEEEPARstart{R}{ecently}, the various forms of advanced driver assistance system (ADAS) for self-driving and autonomous vehicle are receiving a lot of attentions~\cite{shin2019rail,ITS1,pieee202105park,ITS2,ITS3}. For facilitating the various ADAS systems in realistic applications, it is required to form efficient long-term assistance strategies under the consideration of safety according to the fact that the malfunctions in safety can lead to on-road accidents and road congestion. The various ADAS systems developed in modern self-driving and autonomous driving are highly inter-dependency; therefore it does not need to be a single integrated system. Therefore, the strategies that properly and seamlessly control the autonomous vehicle considering various ADAS systems are required.

\begin{figure}[t!]
    \centering
    \includegraphics[width=0.965\columnwidth]{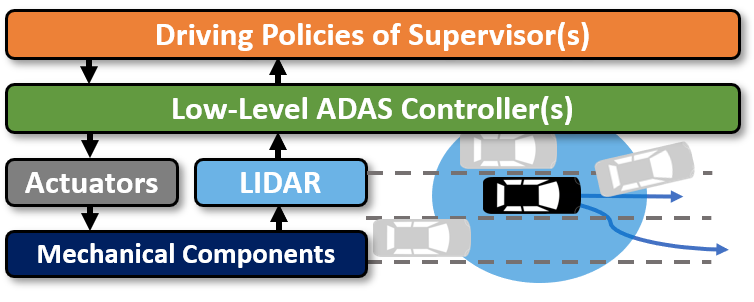}
    \caption{Simplified learning hierarchy to control vehicle systems.}
    \label{fig:overview}
\end{figure}

A self-driving and autonomous vehicle system hierarchy with a supervisor is as presented in Fig.~\ref{fig:overview}. The low-level ADAS controllers are directly connected to LIDAR sensors accessible in the self-driving and autonomous vehicle. The ADAS controllers determine the information needed to efficiently control the vehicle and transmit the determined operations to in-vehicle mechanical components. As a single integrated system, it is expected that multiple ADAS functions simultaneously cooperate to manage the autonomous driving operation. Thus, a supervisor that coordinates the low-level ADAS controllers needs to select suitable ADAS functions when the vehicle operates in on-road driving environments~\cite{korssen2018systematic}. The main objective of the supervisor is decision-making for self-driving and autonomous vehicles during driving operations.

\begin{figure*}[t!]
    \centering
    \includegraphics[width=0.98\textwidth]{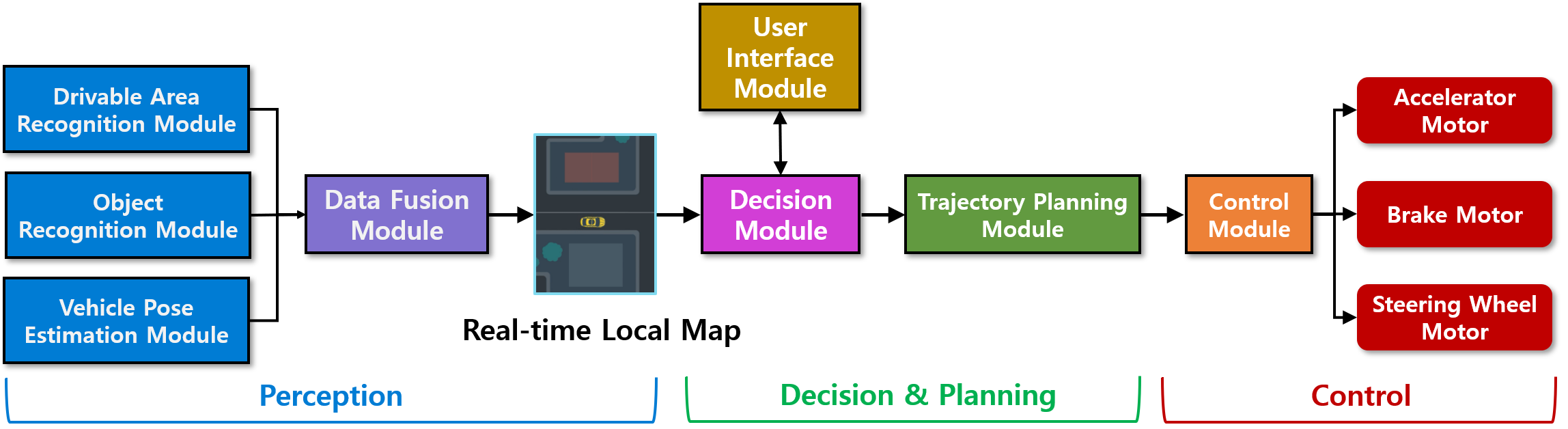}
    \caption{Simplified system hierarchy of autonomous vehicle.}
    \label{fig:systemhierchy}
\end{figure*}

The system architecture of autonomous vehicle is as illustrated in Fig.~\ref{fig:systemhierchy} where the autonomous driving vehicular system is divided into three parts by task, i.e., perception, decision and planning, and control~\cite{zong2018architecture,TASE1}. As a major component of the autonomous vehicle system, the supervisor decides and plans the operation of vehicle in order to realize safety-critical operations. If perception modules cover a sufficient range for autonomous vehicles, the data fusion module generates the high-dimensional observations, e.g., the process of forming the environmental information consisting of object locations, the position and type of path delimiters and the semantic meaning of drivable paths around the vehicle~\cite{pan2017virtual}. Then, the supervisor generates a control plan based on the sensed information by using the decision and planning module. The driver is able to intervene in the operation of the vehicle through the user interface module.

The challenge is that the driving policies of the autonomous vehicular supervisor has to make decisions for the extremely high robustness in any kinds of various traffic environments. One of the major previous research results on self-driving and autonomous driving is rule-based driving policy optimization. However, the policy has serious problems for coping with time-varying environments, i.e., extremely large observation spaces and action spaces which can introduce high computational complexity for training~\cite{gipps1986model,ahmed1999modeling,isj202003saad,isj202103kim,
iot2020kwon,baek2021joint}. Recently, deep reinforcement learning (DRL) based algorithms have been proposed that utilize powerful function approximations such as neural networks, allow the vehicular supervisor to train robust driving policies~\cite{tvt202106jung,mnih2015human,9647911,silver2016mastering,sallab2016end,hoel2018automated,mukadam2017tactical}. However, DRL-based algorithms are also with serious problems when the driving policies conduct policy training that maximizes the expected rewards during operation in real-time. This is due to the fact the determining the reward function values for autonomous driving is still in progress. 
Moreover, due to the fact that there are undesirable policies to maximize the expected rewards while violating the implicit rules of the given driving environments, it is difficult to train the safe and robust policies via DRL-based algorithms in self-driving and autonomous driving~\cite{pan2018agile}.

Based on these problems, many researchers start to consider imitation learning for optimizing the driving policy~\cite{tnnls1}. The imitation learning trains the driving policies based on the desired expert behavior demonstrations rather than the configuration of the reward functions. Furthermore, the imitation learning is able to leverage domain knowledge. Based on these  advantages, it is verified that the imitation learning performs remarkably in many artificial intelligence research areas such as navigation, autonomous vehicle, robotics, and so forth~\cite{pomerleau1991rapidly,pomerleau1989alvinn,pan2018agile}.

However, the main challenge is that the combination of deep reinforcement learning and imitation learning algorithms require huge data for achieving reasonably acceptable performance, and one of the most well-known and famous one is generative adversarial imitation learning (GAIL)~\cite{schulman2015trust,schulman2017proximal}. In order to take care of this problem, the proposed models can be complicated; and therefore, the models introduce reproducibility crisis. Furthermore, the models are sensitive to the implementation of the algorithms and rewards from environments. For instance, in GAIL, the discriminator designed inspired by generative adversarial network (GAN) acts as reward function. With the combination of discrimination and the complex deep reinforcement learning algorithms such as TRPO and PPO, the GAIL learns the its own policies. As a result, the policies based on the reconstruction results do not have always introduce reasonably acceptable performance, and might converge to sub-optimal. These problems introduce the difficulties in training robust and safety-critical autonomous self-driving policies; the trained policies have not been successfully deployed to autonomous vehicles yet~\cite{henderson2017deep,islam2017reproducibility}. Recently, augmented random search (ARS) which is based on the natural gradient policy is proposed~\cite{rajeswaran2017towards,8852307}. According to the fact that the ARS is based on derivative-free simple linear policy optimization, it is relatively easy to reconfigure the robust trained policy that works with reasonably acceptable performance.

In this paper, a novel imitation learning based algorithm is proposed which combines the concepts of ARS and GAIL. 
For more details, a \textit{parallelized and randomized adversarial imitation learning (RAIL)} algorithm is discussed; and the RAIL algorithm trains its own policies via randomly generated matrices where the matrices are used for searching update directions which introduce optimal policies.
This RAIL-based approach is advantageous in computation overhead reduction (in back-propagation) whereas deep reinforcement learning algorithms which utilize gradient descent optimization for computing weight parameters.
Furthermore, our imitation learning based system is able to learn the supervisor's driving policies which achieves almost equivalent performance with the expert in terms of average speeds and lane changes by leveraging expert demonstrations. Based on intensive simulation-based evaluation, it has been demonstrated that the RAIL algorithm is able to train the autonomous self-driving decisions as expected.

\subsection{Contributions}
Our proposed RAIL shows that the random search in the space of policy parameters can be adapted to imitation learning for autonomous driving policies. More details are as follows.
\begin{itemize}
    \item First of all, self-driving mechanism is proposed inspired by imitation learning. Our method can successfully imitate expert demonstrations; and the corresponding policies can achieve similar speeds with lane changes and overtakes.
    \item Next, previous imitation learning methods were based on conventional methods which show complicate configurations to control autonomous driving. However, our proposed RAIL has simplicity based on derivative-free randomized search.
    \item This method has not been previously applied to learn the robust policies in autonomous self-configurable driving.
\end{itemize}

\subsection{Organization}
The rests of this paper are as follows:
Sec.~\ref{sec:related} and Sec.~\ref{sec:background} describe the previous research results and reinforcement learning background knowledge. 
Sec.~\ref{sec:pe} presents problem definition, i.e., policy training for self-driving and autonomous driving. Sec.~\ref{sec:rail} designs the proposed randomized adversarial imitation learning algorithm. Sec.~\ref{sec:experiment} presents the data-intensive simulation-based performance evaluation results in multi-lane highway autonomous vehicle control, and lastly, Sec.~\ref{sec:conclusion} describes the concluding remarks of this paper.

\section{Related Work}\label{sec:related}
\subsection{Imitation Learning}
The imitation learning algorithms can be majorly classified into two categories, i.e., behavioral cloning (BC) and inverse reinforcement learning (IRL). The BC is considered as the simplest imitation learning algorithm. For restoring expert policy, the BC learns by collecting training data from the expert driver’s behaviors, and then the BC utilizes it to directly learn the expert driver's policy. If the policy deviates from expert trajectories which is trained during the training procedure, the imitation learning agent tries to be fragile. This is due to the fact that BC tries to reduce the 1-step deviation error of training, not the error of entire expert trajectories. On the other hand, IRL has an intermediate procedures to estimate the unknown reward function value which numerically represents the expert demonstrations~\cite{ziebart2008maximum,finn2016guided}. According to the fact that the IRL-based algorithms have to train the policy as well as to estimate the reward function values, they generally involve huge computational costs. In~\cite{finn2016connection,ho2016generative}, the theoretical and practical considerations of connections between IRL and adversarial network is studied. GAIL trains a policy which can imitate expert demonstration using the discriminator neural network, which bypasses the reward function value optimization. 

\subsection{The Simplest Model-Free Reinforcement Learning}
The simplest model-free reinforcement learning algorithm which is able to solve standard benchmarks of reinforcement learning is studied with two research directions, i.e., (i) linear policies using natural policy gradients~\cite{rajeswaran2017towards} and (ii) derivative-free policy optimization~\cite{es}. In~\cite{rajeswaran2017towards}, it has been verified that the complicated structures of policies are not needed for solving continuous action control problems. The proposed method in~\cite{rajeswaran2017towards} trains the linear policies using natural policy gradients. The trained policies in~\cite{rajeswaran2017towards} achieves superior performance on complex environments. On the other hand, the proposed method in \cite{es} verify that evolution strategies (ES) guarantees less data efficiency than traditional reinforcement learning algorithms whereas it can provide several advantages. Especially, a derivative-free policy optimization allows ES to be more efficient in distributed and parallized learning environments. Furthermore, the trained reinforcement learning policies tend to be more diverse than the policies obtained by conventional reinforcement learning algorithms. As presented in \cite{ars}, the connection between \cite{es} and \cite{rajeswaran2017towards} is studied for computing the simplest model-free reinforcement learning algorithm which is the derivative-free policy optimization for training linear policies. The proposed simple randomized search method shows state-of-the-art sample efficiency compared to competing learning methods as shown in the simulation results in MuJoCo locomotion benchmark environments.

\section{Background}\label{sec:background}
\subsection{Markov Decision Process}
The Markov decision process (MDP) can be formed as $(S, A, p(s), p(s'|s,a), r(s,a,s'), \gamma)$ where $S$ and $A$ stand for the sets of states and actions, respectively~\cite{twc201912choi,sutton1998introduction,sutton1999between}. Here, $\gamma$ represents the reflection rate of future rewards compared to the current decision. $p(s)$ means initial state probability distribution, in each $s\in S$. $p(s'|s,a)$ stands for the environmental dynamics represented as conditional state distribution in state $s'\in S$ when $s\in S$ (states) and $a\in A$ (actions) are given, i.e., it is defined as the probability for being at $s'$ with action $a$ from $s$. Lastly, $r(s, a, s')$ stands for reward functions for being at $s'$ with action $a$ from $s$. 
The reward returns are defined as $R_t = \sum_{i=t}^{\infty}{\gamma^{i-t}r(s_i,a_i,s_{i+1})}$. The objective of MDP is for finding a policy that maximizes the expected reward returns.

\subsection{Behavior Cloning}
The BC trains a policy as a way of supervised learning over state-action pairs from expert demonstration~\cite{michie1990cognitive}. The objective of BC is as: 
\begin{multline}
    \argmin_{\theta} \mathbb{E}_{s\thicksim P_E}\left[ \mathcal{L}\left( a_E, \pi_\theta\left(s\right)\right)\right] = \\ \mathbb{E}_{s\thicksim P_E}\left[ \left( a_E - \pi_\theta\left(s\right)\right)^2\right]
\end{multline}
where $P_E$ and $a_E$ denote the distributions of the states where expert has been visited and the action of the expert at state $s$ respectively. Note that $\pi_E$ and $\pi_\theta$ represent the policies of the expert and the agent.
According to the fact that BC pursues to minimize 1-step deviation error as a way of supervised learning, the BC is able to train the robust policy when the large amounts of expert demonstration data are provided. Due to this limitation, the trained policy can be fragile when distribution mismatch occurs between the training data and the test data. Therefore, in many research, BC is only used to initialize policy parameters in the absence of expert data. After that, the other algorithms are used to train the robust imitation policy.

\subsection{Inverse Reinforcement Learning}
The IRL can train a policy when MDP formulation is known whereas the reward function values are unknown and expert demonstration data $\mathcal{T}_E$ can be utilized~\cite{ng2000algorithms}. The IRL reveals the hidden reward function values $R^{*}$ that can represent the expert demonstration. 
\begin{equation}
    \label{irl_reward}
    \mathbb{E}\left[\sum_{t=0}^{\infty}\nolimits\gamma^tR^*(s_t) | \pi_E \right] \ge \mathbb{E}\left[\sum_{t=0}^{\infty}\nolimits\gamma^tR^*(s_t) | \pi_\theta \right]. 
\end{equation}
Based on the revealed reward function $R^{*}$, the reinforcement learning is now able to conduct to train the policy $\pi_{\theta}$. Note that the objective of IRL is as:
\begin{equation}
    \label{irl_rl}
    \argmax_{\theta} \mathbb{E}_{s\thicksim P_E}\left[ R^*(s, \pi_\theta(s))\right]. 
\end{equation}
The IRL computes the reward function $R^*$ which represents expert demonstration trajectories. Thus, we do not need to worry about the case where the trained policy to be fragile due to the mismatch between the training data and the testing data, anymore. However, the IRL is computationally expensive to be executed due to the fact that it computes both reward function optimization (\ref{irl_reward}) and policy optimization (\ref{irl_rl}) at the same time. 

\subsection{Generative Adversarial Imitation Learning}
The GAIL can be utilized for reward function definition~\cite{ho2016generative}. Based on GAN, the GAIL trains a discriminator which is a binary classifier, i.e., $D(s,a)$, in order to classify the transitions sampled from an expert demonstration and the transitions sampled by trained policies. The objective of GAIL is as: 
  \begin{multline}
  \label{eq:lossGAIL}
     \argmin_\theta \argmax_\phi \left\{\mathbb{E}_{\pi_\theta} \left[ \log \mathcal{D}_\phi(s, a)\right] +\right. \\ \left. \mathbb{E}_{\pi_E} \left[ \log(1-\mathcal{D}_\phi(s,a))\right]
     - \lambda H(\pi_\theta)\right\}
  \end{multline}
where $\mathcal{D}_\phi(s, a) \to \left[ 0, 1 \right]$ is the discriminator parameterized by $\phi$~\cite{ho2016generative}.
In \eqref{eq:lossGAIL}, $H(\pi_\theta)\triangleq\mathbb{E}_{\pi}\left[-\log\pi(a|s)\right]$ is entropy regularization. In GAIL, the policy receives a reward from the discriminator; and then the policy tries to confuse the discriminator. In this procedure, the reward is optimized using on-policy reinforcement learning optimization schemes. That is, the $\mathcal{D}_\phi$ acts as a reward function in MDP and it gives learning signal to the policy~\cite{ho2016generative,gasil,optiongan}.

\subsection{Augmented Random Search}
The ARS is a model-free reinforcement learning algorithm~\cite{ars}. Based on randomized search in the parameter spaces of policies, the ARS executes the method of finite differences to control and adjust its weights and train the way how the policy performs its given tasks~\cite{matyas1965random,ars}. Via the random search in the parameter spaces, the ARS algorithm conducts a derivative-free policy optimization with noises~\cite{matyas1965random,ars}. For updating the training weights in an effective way, the ARS (i)  uniformly selects update directions and (ii) updates the policies based on the selected direction. 
In order to update the parameterized policy $\pi_\theta$, the update directions can be $
    \frac{r(\pi_{\theta - \nu\delta}) - r(\pi_{\theta + \nu\delta})}{\nu}$ where $\delta$ is a $0$ stands for a Gaussian vector, $\nu$ is a positive real number that represents the standard deviation of exploration noise, and $r(\pi_\theta \pm \nu \delta)$ is the reward from environments when the parameter of policies is given as $\pi_\theta \pm \nu\delta$.
Here, $\theta_t$ is the weight of policy training at $t$-th iteration. $N$ denotes the number of sampled directions per iteration. The update step is as:
\begin{equation}
    \label{brs}
    \theta_{t+1} = \theta_{t} + \frac{\alpha}{N} \sum_{i=1}^{N}\nolimits{\left[r(\pi_{\theta + \nu\delta_i}) - r(\pi_{\theta - \nu\delta_i})\right]\delta_i}.
\end{equation}

However, this randomized search is challenging due to  the large variations in terms of the rewards $r(\pi_\theta \pm \nu\delta)$ those are observed during training procedure. The variations is harmful doe because the updated policies become to be perturbed during the update steps~\cite{ars}. In order to handle the large variation, the standard deviation of the rewards that is collected at each iteration, i.e., $\sigma_R$, is is utilized to control and adjust the sizes of the update steps during ARS. Based on this concept for step size adaptation, the ARS performs better compared to the other conventional deep reinforcement learning algorithms, e.g., PPO, TRPO, in specific learning environments.

\section{Problem Definition}\label{sec:pe}
\subsection{Motivation}
For coordinating ADAS functions for safe and robust autonomous driving control, the vehicular supervisor determines the optimal ADAS functions based on the nearby environments and situations. However, the complete environment states are unknown to the supervisor. 
The vehicular supervisor gather an observation which is conditioned on the current state. The host vehicle interacts with the high-way environment including nearby multiple vehicles and lanes; in addition, it utilizes partially observable local information. Therefore, the observation of agent is able to be modeled as $(O, A, T, R, \gamma)$ representing a partially observable MDP with continuous observations and actions. Similar to MDP, there are partial observation states denoted by $O$ instead of $S$. Note that LIDAR sensing data is considered as the observation. 

Two types of spaces are considered in this paper, i.e., a finite state space $\mathcal{O} \in \mathbb{R}^n$ and a finite action space $\mathcal{A} \in \mathbb{R}^p$. The objective of imitation learning is to train a policy $\pi_\theta  \in \Pi : \mathcal{O} \times \mathcal{A} \to \mathbb{R}^p$ that imitates expert demonstration by GAN $\mathcal{D}_\phi(s, a) \to \left[ 0, 1 \right]$ where $\theta \in \mathbb{R}^n$ and $\phi \in \mathbb{R}^{n+p}$ are the policy parameters and the discriminator parameters, respectively~\cite{ho2016generative}.

\subsection{State Space} 
For LIDAR sensing data model, a vector observation is used. In this model, $N$ beams are spread evenly over the field of view $\left[ \omega_{min}, \omega_{max} \right]$. Each LIDAR sensor data has a maximum range of $r_{max}$ surrounding the host vehicle. Then, the observation is described as $O = (o_1,\dots,o_N )$. Furthermore, the speed of the obstacle and the speed of host vehicle can be obtained based on the distance information. Note that the speed observation is denoted by $V_r = (v_1,\dots,v_N)$.

\subsection{Action Space}
The policy in a supervisor is considered as a high-level driving decision maker that determines optimal actions based on surrounding observation on the highway. According to the fact that the self-driving and autonomous vehicle utilizes the equipped multiple ADAS functions, the determined actions of autonomous driving policy utilizes each ADAS function. Note that the driving policy is defined in a high level five decisions (i.e., discrete five action spaces), i.e., maintaining current status, accelerating speed as $vel_{cur} + vel_{acc}$, decelerating speed as $vel_{cur} - vel_{dec}$, making a left lane change, and making a right lane change. This is based on the assumption that the host vehicle can be perfectly controlled with autonomous emergency braking and adaptive cruise control~\cite{mukadam2017tactical,min2018deep,hoel2018automated}. 

\subsection{Reward Function}
In GAIL, the reward is defined as $r_{\pi_\theta}(s, a) = - \log(1 - \mathcal{D}_\phi(s, a))$ or $r_{\pi_\theta}(s, a) = \log(\mathcal{D}_\phi(s,a))$~\cite{ho2016generative} where the first reward is considered to let an agent train survival policies via a survival bonus with the positive reward form based on their lifetime whereas the second one is considered to train policies with a per-step negative reward, when the reward consists of the negative constant for states and actions. However, in this case, learning the survival policies are not easy~\cite{kostrikov2018discriminator}. The prior knowledge of environmental objectives is worthy, however the environment-dependent reward is undesirable when an agent requires interactions with a training environment for imitating an expert policy. Thus, the reward function is defined as:
\begin{multline}
    r(s ,a) = \\ \mathbb{E}_{(s, a)\thicksim \pi_\theta}\left[\log(\mathcal{D}_\phi(s,a))- \log(1 - \mathcal{D}_\phi(s,a))\right].
\end{multline}

\section{Randomized Adversarial Imitation Learning}\label{sec:rail}
\vspace{-0.1mm}

\begin{figure*}[t!]
    \centering
    \includegraphics[width=1\textwidth]{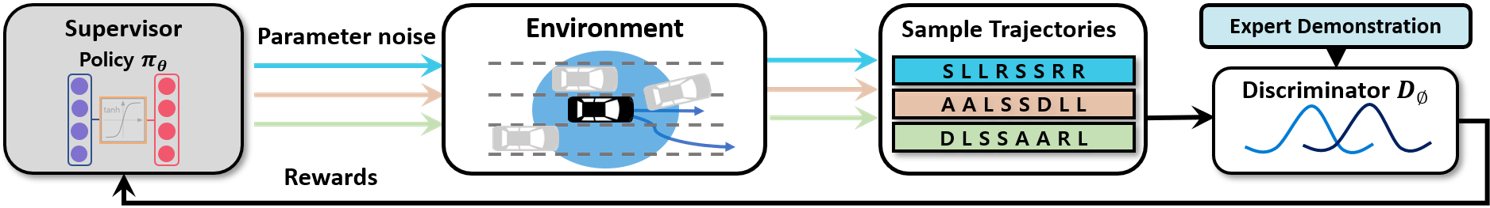}
    \caption{Structure of RAIL.}
    \label{fig:structure}
\end{figure*}

\subsection{Algorithm}
This paper proposes a \textit{randomized adversarial imitation learning (RAIL)} which adopts imitation learning through adversarial network paradigm, i.e., generative adversarial imitation learning (GAIL). 
The proposed RAIL algorithm aims to train the driving policy $\pi_\theta$ for imitating expert driving demonstration. This section describes the details of the proposed RAIL algorithm and makes a connection between GAIL and ARS where the ARS is based on a derivative-free optimization.  

As presented in Fig.~\ref{fig:structure}, the entire architecture of RAIL algorithm is illustrated. The supervisor on top of our considering host vehicle is an agent with the policy of $\pi_\theta$. Based on the given environment (i.e., multi-lane highway environment), our considering host vehicle gathers the observation at first. After that, the random noise matrices with small values are generated where the noise element values are added to or subtracted from the driving policy parameters $\theta$. As a result, temporal several different policies are generated. The agent/policy interacts with the given driving environments multiple times based on the generated noisy policies and the results are gathered as sample trajectories. Based on the sample trajectories, the agent policy $\pi_\theta$ is trained in order to conduct autonomous vehicle control successfully with the ADAS functions that guarantee driver/vehicle safety. 
During the policy training process, the policy $\pi_\theta$ tries to disturb the discriminator $\mathcal{D}_\phi$ by let the discriminator believe the sample agent trajectories obtained by expert demonstrations. The $\mathcal{D}_\phi$ pursues to classify between the distribution of trajectories those are sampled by the policies $\pi_\theta$ and the expert trajectories $\mathcal{T}_E$ where the trajectories are formulated as state-action pairs $(s, a)$. The discriminator acts as the reward module in RAIL, as presented in Fig. \ref{fig:structure}; and therefore $\pi_\theta$ is trained against the discriminator.

As presented in Fig.\ref{fig:structure}, the discriminator conducts its own training with sample trajectories and expert demonstration. During the training procedure, the distribution of the sample trajectories changes because the policy $\pi_\theta$ is updated every iteration. As a result, the training is not stabilized; and therefore it leads to the inaccurate reward signal to the discriminator policy. In turn, the policy is perturbated during updates~\cite{gasil}. 
In RAIL, the loss function of least square GAN (LS-GAN) is utilized for training the discriminator policy $\mathcal{D}_\phi$~\cite{mao2017least}, and the objective is as:
\begin{equation}
  \label{lsgan_loss}
  \begin{multlined}
     \argmin_\phi L_{LS}(\mathcal{D}) = \frac{1}{2}\mathbb{E}_{\pi_E} \left[(\mathcal{D}_\phi(s, a) - b)^2\right]+\\
     \frac{1}{2}\mathbb{E}_{\pi_\theta} \left[ (\mathcal{D}_\phi(s,a) - a)^2\right]
\end{multlined}
\end{equation}
where $a$ is the discriminator label for the sampled trajectories from the policy $\pi_\theta$ 
and $b$ is the discriminator labels for the expert trajectories, respectively. 

As mentioned, LS-GAN loss function is utilized for training the discriminator policy. When the loss function of conventional GAN in (\ref{eq:lossGAIL}) is used, sampled trajectories those are far from the expert trajectories whereas the correct side of the decision boundary are almost not penalized by sigmoid cross-entropy loss function. On the other hand, the LS-GAN loss function (\ref{lsgan_loss}) penalizes the sampled trajectories those are far from the expert trajectories on either side of decision boundary~\cite{mao2017least}. Thus, the stability of training can be improved; and it also leads the discriminator to provide accurate reward signals to update step. In LS-GAN, $a$ and $b$ are in the relationship of $b-a=2$ for \eqref{lsgan_loss} to be Pearson $\mathcal{X}^2$ divergence~\cite{mao2017least}. However, $a=0$ and $b=1$ are used for the target discriminator labels. The results of the discriminator $\mathcal{D}_\phi$ are between 0 and 1; and these values are experimentally obtained. In RAIL, the discriminator is interpreted as a reward function for policy optimization. Forementioned in Sec.\ref{sec:pe}, the form of reward signal is as:
\begin{equation}
  \label{reward_signal}
  \begin{multlined}
     r_{\pi_\theta}(s, a) = \log(\mathcal{D}_\phi(s,a))- \log(1 - \mathcal{D}_\phi(s, a)).
\end{multlined}
\end{equation}

This means the policy $\pi_\theta$ gets higher reward $r_{\pi_\theta}(s, a)$ when the trajectories sampled from the policy $\pi_\theta$ is almost equivalent to expert trajectories. The $\pi_\theta$ is updated to optimize/maximize the discounted summation of reward values given by the discriminator rather than the reward values from the environment as in Fig.~\ref{fig:structure}. The objective of RAIL is as \begin{equation}
\argmax_{\theta} \mathbb{E}_{(s, a)\thicksim \pi_\theta}\left[ r(s, a)\right], 
\end{equation}
and by \eqref{reward_signal},
\begin{equation}
  \label{RAIL_objective}
    \argmax_{\theta}    \mathbb{E}_{(s, a)\thicksim \pi_\theta}\left[ \log(\mathcal{D}_\phi(s,a))- \log(1 - \mathcal{D}_\phi(s, a)) \right]
\end{equation}
and this represents the connection between randomized parameter space search in RAIL and adversarial imitation learning.

\begin{algorithm}[t]
    \textbf{Hyperparameters}\\
    $\alpha$: learning rate, 
    \\ $N$: the number of directions per iteration,
    \\ $\delta$: Gaussian noise vector, 
    \\ $\nu$: noise coefficient,
    \\ $\eta$: evaluation term, 
    \\ $\tau$: noise effect coefficient, \\
    $\theta$: weight vectors \\
    \textbf{Initialization}\\
    $\mu_0 \leftarrow \bold{0}\in \mathbb{R}^n,$ and $\sum_{0} \leftarrow  \bold{I}_n \in \mathbb{R}^{n\times n}$\\
    \While{$t \le$ episode}
    {
        \textbf{Random Sampling (i.i.d.)} $\bold{\delta_t} \triangleq \left\{ \delta_1, \delta_2, . . . , \delta_N ; \delta^i_k \in \mathbb{R}^{h \times n}, \delta^o_k \in \mathbb{R}^{p \times h} \right\} $ \\
        \textbf{Collect rollouts and rewards with noisy policies for} $k\in \left\{ 1,2,\dots,N \right\}$:\\
        \Indp
        $\pi_{t,(k),+} \leftarrow (\theta_t + \nu\delta_k)$diag$(\sum_t)^{-1/2}(s - \mu_t )$\\
        $\pi_{t,(k),-} \leftarrow (\theta_t - \nu\delta_k)$diag$(\sum_t)^{-1/2}(s - \mu_t )$\\
        \Indm
        
        \textbf{Update the discriminator param. $\phi_t$}:\\
        \Indp
        $\nabla_{\phi_t}L_{LS} \leftarrow \frac{1}{2}\mathbb{E}_{\pi_E} \left[(\nabla_{\phi_t}\mathcal{D}_{\phi_{t}}(s, a) - b)^2\right]$\\ 
        \Indp
        \Indp
        \Indp
        \Indp
        $+\frac{1}{2}\mathbb{E}_{\pi_\theta}\left[(\nabla_{\phi_{t}}\mathcal{D}_{\phi_{t}}(s,a) - a)^2\right]$\\
        \Indm
        \Indm
        \Indm
        \Indm
        \Indm
        \textbf{Update the policy param. $\theta_t$}:\\
        
        $\theta_{t+1} \leftarrow \theta_{t} + \frac{\alpha}{N\sigma_R}\sum^{N}_{i=1}{\left[ r(\pi_{t,(k),+}) - r(\pi_{t,(k),-}) \right]\delta_{(k)} }$\\
        \Indp
        where reward is 
        \Indm
        $r(\pi_{t,(k),\pm}) \leftarrow \mathbb{E}_{(s,a)\thicksim \pi_{t,(k),\pm}} \left[ \log(\mathcal{D}_\phi(s,a))- \log(1 - \mathcal{D}_\phi(s, a)) \right]$.\\
        
        Set $\mu_{t+1}$ (mean) and $\sum_{t+1}$ (co-variance) of the states encountered by training starting.\\
        \eIf{t == $\eta$ and Performance is not improved}{
            $\nu$ += $\tau$\;
        }
        {
            $\nu$ = $\nu_{init}$
        }
        $t$ = $t + 1$
    }
    \caption{RAIL}
    \label{algo:RAIL}
\end{algorithm}

The proposed RAIL algorithm is related to ARS that is a kind of model-free reinforcement learning. Therefore, the RAIL algorithm utilizes parameter space exploration via derivative-free policy optimization. In this paper, $\pi_\theta$ are denoted by $\theta$ where the $\pi_\theta$ consists of $\pi^i_\theta$, $\pi^o_\theta$, and activation. Note that the input layer of $\pi_\theta$ is denoted by $\pi^i_{\theta^i}$ where $\theta^i \in \mathbb{R}^{n\times h}$ are the parameters of the layer. Similarly, the output layer is denoted by $\pi^o_{\theta^o}$ where $\theta^o \in \mathbb{R}^{h\times p}$. The noises of parameter space, i.e., $\delta^i$ and $\delta^o$, are formed as the matrices of $n \times h$ and $h \times p$, respectively, where the values are randomly sampled with $0$ mean and $\nu$ standard deviation in Gaussian normal distribution. Note that $\theta$ is a set of $\theta^i$ and $\theta^o$; and $\delta$ is a set of $\delta^i$ and $\delta^o$.

The proposed RAIL algorithm works as presented in Algorithm~\ref{algo:RAIL} where the parameters, i.e.,  $\theta^i$ and $\theta^o$, are initialized by BC. During the training procedure, the noises are chosen randomly for each iteration where they are denoted by $\bold{\delta^i}$ and $\bold{\delta^o}$ which stand for the policy search directions in parameter space (line 9). Each set of selected $N$ noises let two policies be $\pi_\theta$ (i.e., current policy). $2N$ rollouts and rewards are collected by $N$ noisy policies $\pi_{t,k,\pm} = \theta_{t} \pm \nu\delta_k$ (line 10-12). The state normalization is used in RAIL (line 6, 19); and it makes policy $\pi_{t,i,\pm}$ have equal influence for the changes of state components when there are state components with various ranges~\cite{ars,es,nagabandi2018neural}. 
Notice that the reason why state normalization is required is that the high dimensional problems have multiple state components with various ranges; and therefore it makes the policies to result in large changes in actions when the equivalent sized changes are not equally influence states.
The discriminator $\mathcal{D}_\phi$ gives the reward to update steps. However, because the trajectories for the discriminator training are only able to be obtained by $\pi_{\theta_t}$ (current policies), the discriminator is trained whenever $\theta_t$ is updated. The discriminator $\mathcal{D}_\phi$ computes $\phi$ that minimizes (\ref{lsgan_loss}) (line 14-15). By using the reward by the discriminator, the policy of the neural network weight is updated toward the direction of $+\delta$ or $-\delta$ on top of the result of $r(\pi_{t,(k),+}) - r(\pi_{t,(k),-})$ (line 17-18). The state normalization is by the states encountered during the policy training; and therefore $\mu$ and $\sum$ are updated (line 19). The noise coefficient $\nu$ is initialized as a small real number. Then, the fixed coefficient $\nu$ that is too small can make the agent never converge or get trapped in a suboptimal solution. Therefore, we periodically check the performance of the current agent and increase the noise coefficient $\nu$ whenever we detect performance saturation (line 20-24).

\subsection{Parallization}
\begin{figure*}[t!]
    \includegraphics[width=1\textwidth]{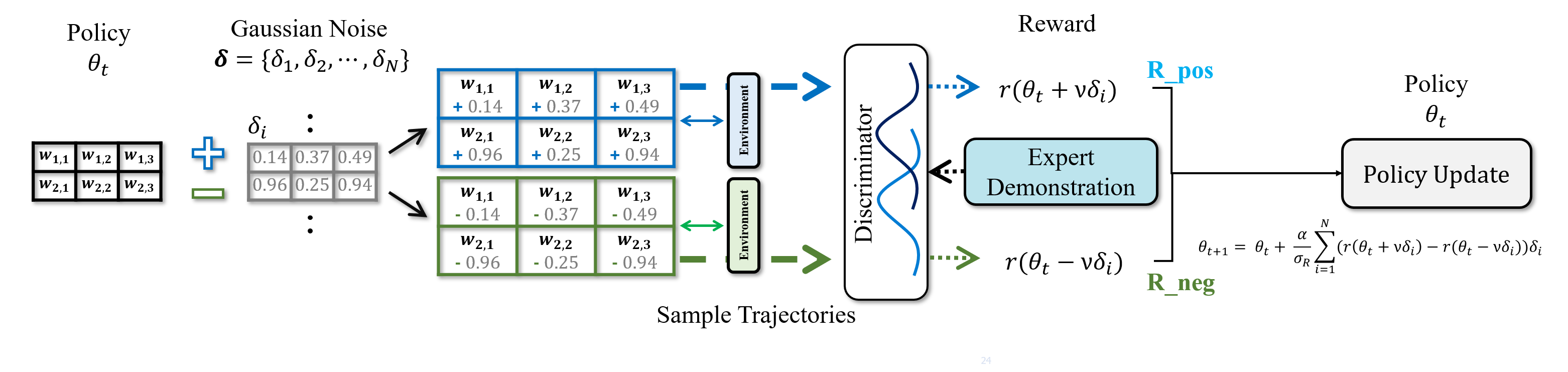}
\caption{Parallization}
\label{fig:parallel1}
\end{figure*}

This section introduces the parallelization of RAIL by using sampled directions $N$. Fig.~\ref{fig:parallel1} shows the overall policy update procedure of RAIL. During the training process, the policy has the matrix of weights. At time $t$, RAIL takes a Gaussian noise vector in the form of weights. At the same time, we generate $N$ number of directions. Each noise vector is then sent to the process separately, and then each process performs the following procedure. Each process adds a noise vector $\delta_t$ to the weights $\theta_t$. At the same time, each process subtracts a noise vector $\delta_t$ from the weights $\theta_t$. Then, each process has two matrices of noise weights, which are called perturbation matrices. The perturbation matrices get it’s own episode. Each process can get different rewards as a result of episode. Each process receives the rewards of episode from the discriminator. The rewards mean how similar the state and action sequences of perturbation matrices are to the expert demonstration. Then, each process calculates $r(\theta_t + \nu\delta_t) - r(\theta_t - \nu\delta_t)$. The result is sent to the main process that manages the update procedure and propagates the updated policy to each process. The main process collects these values and updates the policy.

\section{Experiment-based Performance Evaluation}\label{sec:experiment}
\begin{figure*}[t]
\begin{subfigure}{.5\textwidth}
  \centering
  \includegraphics[width=1\columnwidth]{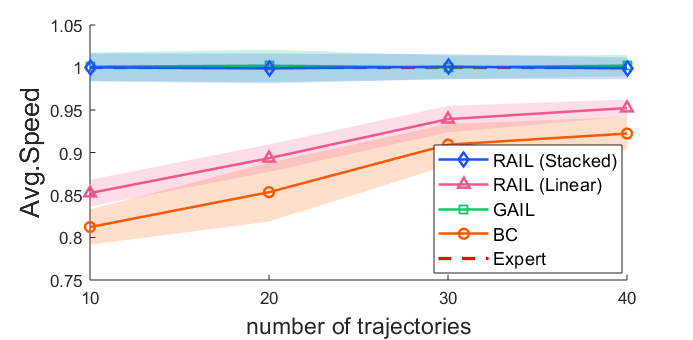} 
  \caption{Speed (Normalized).}
  \label{fig:speed}
\end{subfigure}
\begin{subfigure}{.5\textwidth}
  \centering
  \includegraphics[width=1\columnwidth]{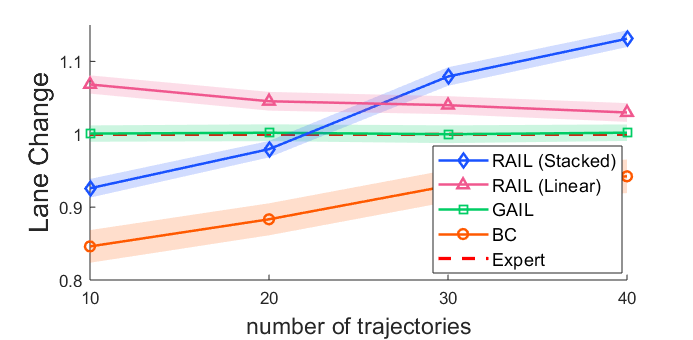} 
  \caption{Lane Change (Normalized).}
  \label{fig:lane}
\end{subfigure}

\begin{subfigure}{.5\textwidth}
  \centering
  \includegraphics[width=1\columnwidth]{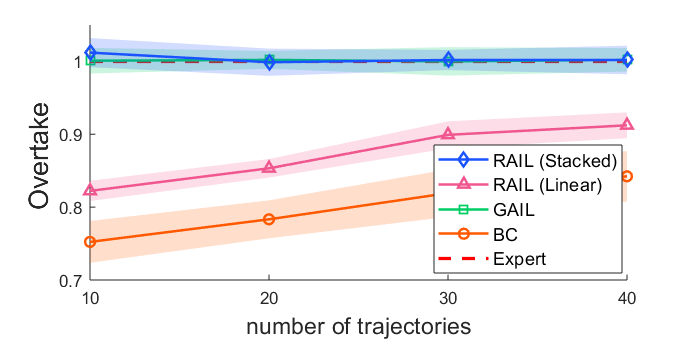} 
  \caption{Overtake (Normalized).}
  \label{fig:over}
\end{subfigure}
\begin{subfigure}{.5\textwidth}
  \centering
  \includegraphics[width=1\columnwidth]{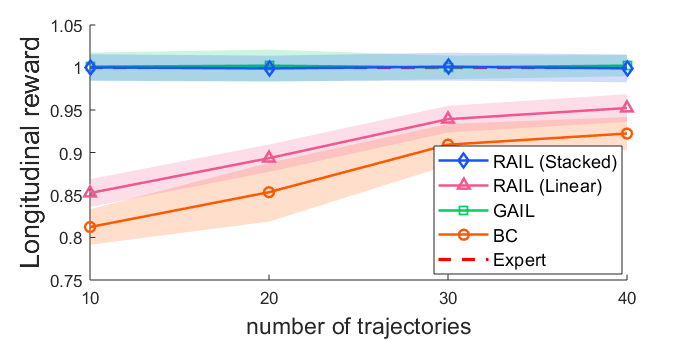}  
  \caption{Longitudinal Rewards (Normalized).}
  \label{fig:long}
\end{subfigure}

\begin{subfigure}{.5\textwidth}
  \centering
  \includegraphics[width=1\columnwidth]{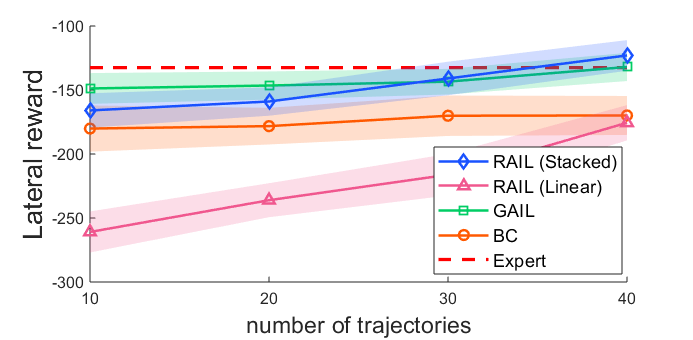} 
  \caption{Lateral Rewards.}
  \label{fig:later}
\end{subfigure}
\begin{subfigure}{.5\textwidth}
  \centering
  \small
    \begin{tabular}{l|c|c|c}
    \toprule
        Average & Expert & RAIL & RAIL\\
            & & (Stacked) & (Linear) \\
    \midrule [1.0pt] 
        Speed &  68.83\,km/h & 70.38\,km/h & 65.00\,km/h \\
        Number of overtake & 44.48  & 45.04  & 40.03 \\
        Number of lane change & 14.04  & 15.01 & 13.05 \\
        Longitudinal & 2642.11  & 2719.38 & 2495.57\\
        Lateral  & -132.52 & -122.98 & -175.6\\[0.2ex]
        \bottomrule
    \end{tabular}
  \caption{Performance (Average: 16 Episodes, 40 Trajectories)}
  \label{tab:experiment}
\end{subfigure}
\caption{The performance evaluation results of trained policy depending on the number of expert trajectories (Average: $5$ episodes).}
\label{fig:sample}
\end{figure*}

This section presents the various simulation results we utilized to verify the performance of the proposed derivative-free imitation learning in autonomous vehicles with various ADAS functions. The performance is evaluated with the simulator by using the results of the corresponding experiments with respect to the aforementioned rewards, given that the state of the autonomous vehicle is observable by the agent. In section~\ref{sec:layer}, the performance between RAIL and baselines is compared. For assessing the performance gaps between the single-layer policy and multi-layer policy trained by RAIL, the single-layer and two-layer policies are implemented. In section~\ref{sec:parallel}, we study the impact of the number of sampled directions per iteration $N$ in terms of reward from discriminator.

\subsection{Simulation Settings}\label{sec:simulator}
In this section, the implementation details of the RAIL algorithm for autonomous vehicle with various ADAS functions are introduced. The hardware configuration we employed for our simulation is an NVIDIA DGX personal AI supercomputer with 4 $\times$ Tesla V100 GPUs and Intel Xeon E5--2698 v4 (2.2 GHz CPU with 20 cores). Python version 3.6 on Ubuntu 16.04 LTS is used for software systems.
During the training procedure, the learning rate was set to $\alpha = 0.001$. The number of directions per iteration was set to $N = 512$ in subsection~\ref{sec:layer}. The noise effect coefficient was set to $\tau = 0.001$. The initial noise coefficient was set to $\nu = 0.03$.
Note that the considering road environment in this paper is a highway driving roadway consisting of five lanes. Aforementioned in Sec.~\ref{sec:pe}, the observation is based on LIDAR sensing data. We assume that the LIDAR sensors detect $360$ degree ranges where one ray is taking care of $15$ degree. The ray returns the distance between the first obstacle it encounters and the main RAIL algorithm-equipped vehicle. If no obstacles exist, the maximum sensing range values are returned. The expert demonstration is designed inspired by PPO. The performance evaluation results show the average of 16 experimental results. In the experiments, the trained weights by BC are utilized for fast convergence in the both of GAIL and RAIL. This simulation design is mainly based on \cite{min2018deep}. Unity was used to implement the multi-lane highway simulator.

\subsection{Performance Comparison by the Number of Layers}\label{sec:layer}
In this paper, RAIL trains the agent that imitates the behaviors of expert vehicles with various ADAS functions. Therefore, longitudinal rewards and lateral rewards, which represent the operation status of various ADAS functions during the autonomous driving, were investigated for each of following four methods: single layer RAIL, stacked layer RAIL, BC, and GAIL. Furthermore, the average vehicle speeds, the number of vehicle overtakes, and the number of vehicle lane changes during the episodes were evaluated with four different methods. These experimental results show the agent imitates the autonomous vehicle operation with various ADAS functions as well as the decision (i.e., change lane to left when vehicle can change both directions) when it needs to decide the action. Table~\ref{tab:experiment} shows the statistical information of the above experiments for four different agents.

The experiment of Fig.~\ref{fig:over} and Fig.~\ref{fig:lane} was conducted to measure the agent successfully imitates decisions from the behaviors of expert. For achieving the similar statistics of expert overtakes, the vehicle lane change points and change directions have to be similar to expert behaviors in the episode. Similarly, in the experiment of Fig.~\ref{fig:speed}, the acceleration or deceleration decisions should be similar to the expert for achieving the similar average speed in the episode.

Table~\ref{tab:experiment} shows that the two-layer policy achieves the highest values of average speed and average overtakes, i.e., $70.38$\,km/h and $45.04$, respectively. This is due to the fact that the trained policies are able to occasionally achieve superior performance than the expert demonstration because GAIL-based imitation learning frameworks conduct policy optimization based on the interaction with the surrounding environments. On the other hand, the performance of single-layer policy achieves approximately $90\%$ performance compared to the expert demonstration. This is due to the fact that the single-layer policy is not enough to properly take care of high dimensional observations. Aforementioned, BC pursues the minimization of 1-step deviation errors along with the expert demonstration. As a result, the single-layer policy achieves undesirable performance according to the distribution mismatch between testing and training.

In Fig.~\ref{fig:over}, the two-layer policy presents the desired performance. This result is related with the tendency that is presented in Fig.~\ref{fig:speed} and Fig.~\ref{fig:lane}. The two-layer policy changes more lanes than the single-layer policy. The lane changes of two-layer policy can make the agent to get proper decision points for acceleration and lane change; then it leads to the average speeds, the number of vehicle lane changes, and the number of overtakes, to be similar to the number of expert demonstration.

However, single-layer policy presents the smaller number of overtakes than the two-layer. Notice that the number of lane changes of the single-layer policy is more than the one of expert demonstration. The single-layer tries to find the points to acceleration. However, the single-layer policy cannot accelerate the autonomous vehicle properly. This result leads the agent to get a small number of overtake.

The experiments for longitudinal reward were performed to analyze the operation status of various ADAS functions of the agents in episodes. In this work, we assumed that the autonomous vehicle drives with smart cruise control. Then, the smart cruise control system is limited to only controlling the speed of the vehicle in this paper. The smart cruise control makes the decision about the speed. According to the fact that the longitudinal reward value is proportional to the vehicle speed, the operation status of the smart cruise control of the trained agent in terms of the speed are evaluated as shown in Fig.~\ref{fig:speed}.

The two-layer policy achieves more similar speed than the single-layer. This is due to the fact that the lane change decisions with two-layer policy make the agent to get the similar average speed compared with the expert.

In this paper, we assumed that the autonomous vehicle is equipped with lane keeping system. Therefore, the experiments for lateral reward was also performed to analyze the operation status of various ADAS functions similar to the longitudinal reward. Until the lane changes are conducted, the driving vehicle is able to change the lane change decision due to the observable states. Due to the fact that the lateral reward continuously occurs during lane change, the frequent changes in each episode let rewards reduction be reduced.
As presented in Fig.~\ref{fig:later}, two-layer policy achieves more reward than the expert demonstration when we trained the agent with 40 expert trajectories.
The two-layer policy presents more frequent lane changes than the behaviors of expert's driving. It means the two-layer policy keeps the decision more than the expert demonstration during the lane change. This result shows the major characteristic of GAIL-based imitation learning method. This experiment shows that RAIL follows the characteristic of GAIL-based imitation method; and thus it shows the applicability of a derivative-free optimization method to the imitation learning.
The single-layer policy achieves the smallest lateral reward value in all cases. Note that the single-layer policy algorithm presents the frequent lane changes comparing to the number of lane changes by the driving expert. This is due to the fact that the policy changes its decision frequently during the lane change operation.
Therefore, the experiments Fig.~\ref{fig:speed}--\ref{fig:over} show that, with the single-layer policy and the two-layer policy, the stacked layer policy for the autonomous vehicle with various ADAS functions is capable of learning the optimal operation better than the single-layer policy.
In summary, the proposed RAIL algorithm is verified that it improves the average speed and also reduces the number of unnecessary lane changes rather than the behaviors of BC. This explains that the RAIL algorithm successfully imitates driving policies from the expert driving demonstrations to the desired directions. This simulation-based experiment results for performance evaluation verify that the two-layer policy achieves desired performance improvements. The results also show the possibility for imitating autonomous vehicle systems by using the derivative-free imitation learning method, called RAIL.
\subsection{Learning Curves Comparison by the Number of Sampled Directions}\label{sec:parallel}
\begin{figure*}[t!]
\begin{subfigure}{.5\textwidth}
  \centering
    \includegraphics[width=1\columnwidth]{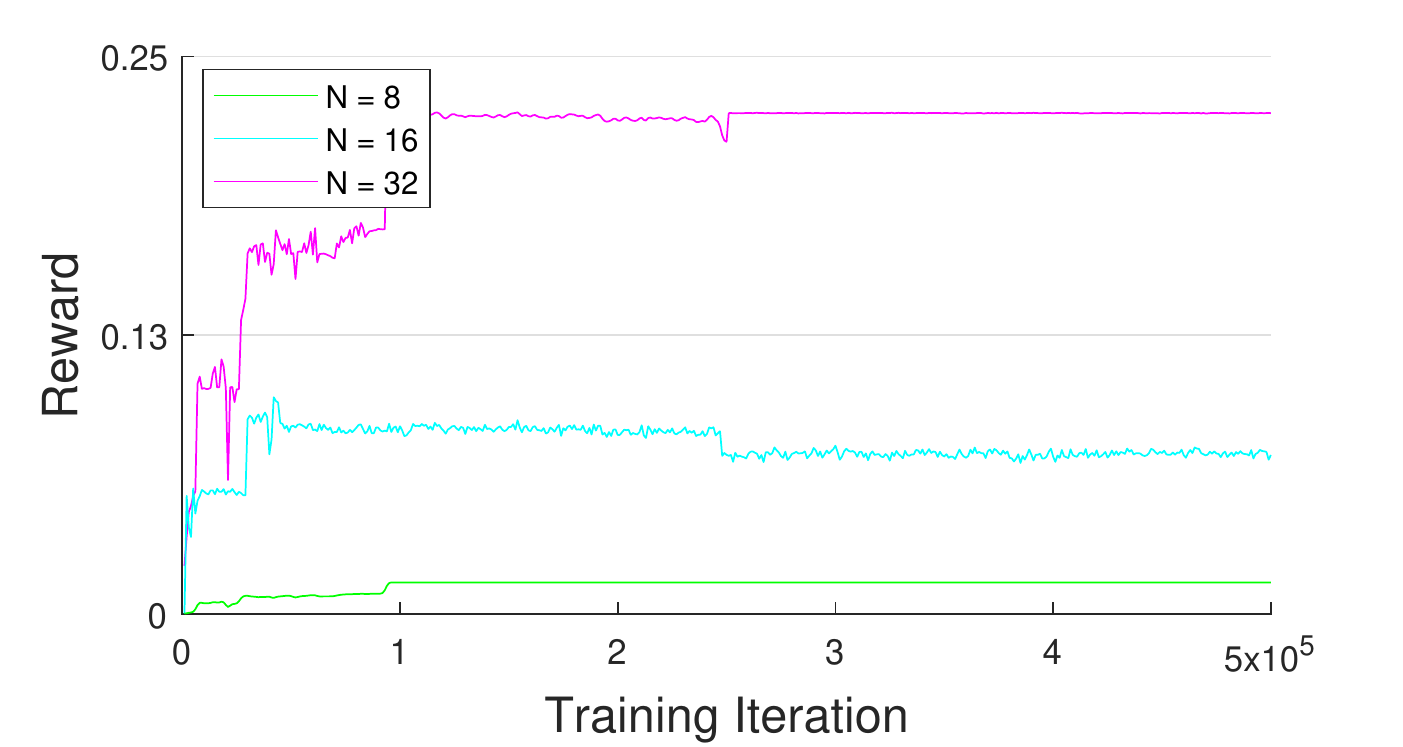}
    \caption{Normalized Rewards (N = 8, 16, 32).}
    \label{fig:reward16}
\end{subfigure}
\begin{subfigure}{.5\textwidth}
  \centering
    \includegraphics[width=1\columnwidth]{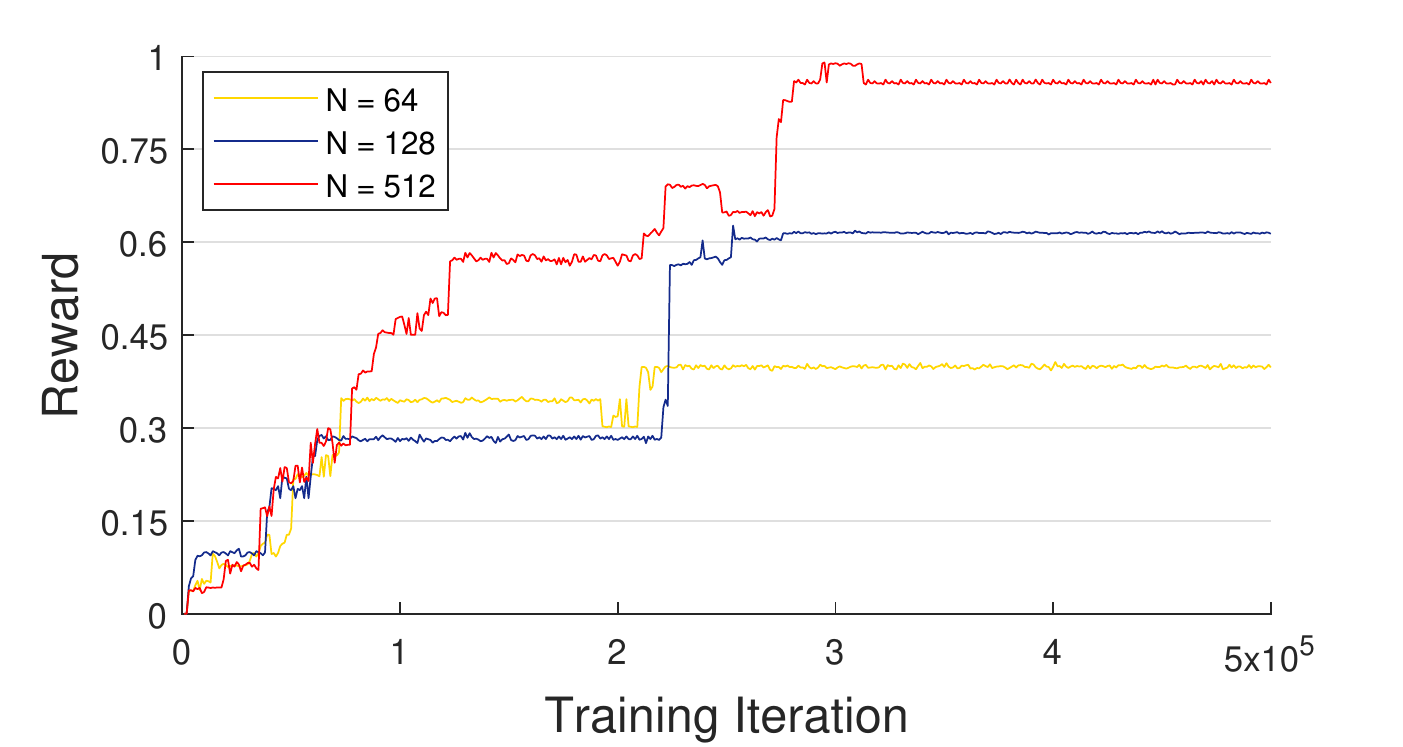}
    \caption{Normalized Rewards (N = 64, 128, 512).}
    \label{fig:reward128}
\end{subfigure}

\begin{subfigure}{.5\textwidth}
  \centering
  \includegraphics[width=1\columnwidth]{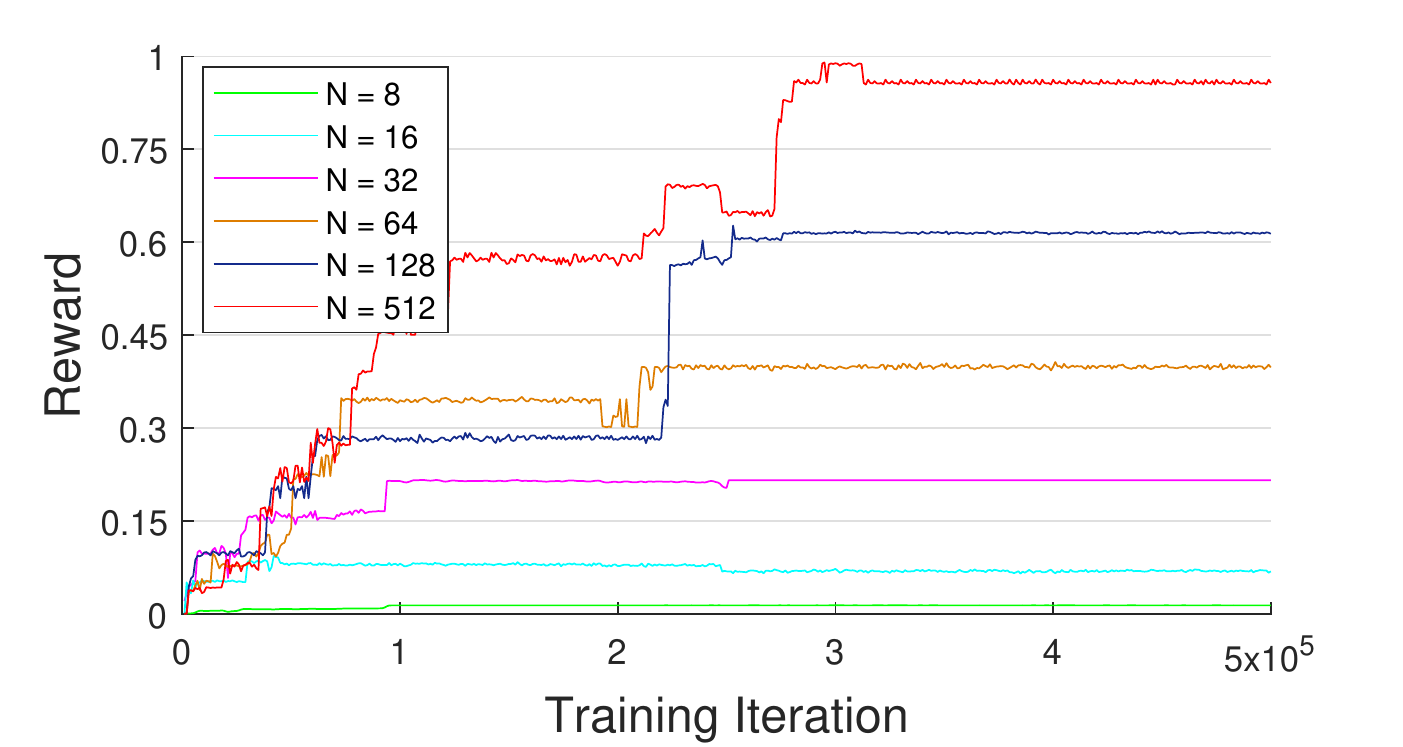} 
  \caption{Normalized Rewards.}
  \label{fig:rewardtot}
\end{subfigure}
\begin{subfigure}{.5\textwidth}
  \centering
  \small
    \begin{tabular}{l|c|c|c|c|c|c}
                    \toprule
                    N & 8 & 16 & 32 & 64 & 128 & 512\\
                    \midrule [1.0pt] 
                    Rewards & 449 & 2278 & 7185 & 13258 & 20453 & 28569 \\[0.2ex]
                    \bottomrule
    \end{tabular}
  \caption{Convergence Rewards from discriminator in episode.}
  \label{tab:experiment}
\end{subfigure}
\caption{The comparison of training curves due to the number of sampled directions in each iteration ($N$).}
\label{fig:parallel}
\end{figure*}

In this subsection, we compare the learning curves of RAIL. The number of sampled directions considerably influence the convergence of the reward throughout the training procedure to train the autonomous vehicle decision maker. Fig.~\ref{fig:parallel} shows the tendency of the reward convergence for six different numbers of sampled directions $N$ in Algorithm~\ref{algo:RAIL}. Note that the results in Fig.~\ref{fig:reward16} to Fig.~\ref{fig:rewardtot} were the results of experiments by using the same settings which is used in the subsection~\ref{sec:layer} with the different number of sampled directions. Note that we did not use the BC-based initialization which is used to accelerate the training procedure. The experiments represent the learning tendency for each reward from the discriminator in one episode. As mentioned before, the discriminator returns the similarity as a reward to the agent during the training procedure. This means that the agent observes a lot of states similar to the expert during the episode.

As shown in Fig.~\ref{fig:reward16}, throughout the training procedure, the performance of the agent is increasing in all cases. When we set to $N = 8$ to train the agent, the training curve increases and then converges. However, the agent trained with $N = 8$ in Fig.~\ref{fig:reward16} shows the poor performance after the reward has converged. Then, the agent has converged to $449$ as shown in Fig~\ref{tab:experiment}. This is because the number of sampled directions $N = 8$ is not enough to get the optimal update directions. On the other hand, when we set to $N = 16$, the training curve quickly increases and converges to the better performance compared to the case of $N = 8$. The agent has converged to $2278$. However, the agent has also converged to the poor performance $7185$. When $N = 32$, the training curve increases rapidly, stalls briefly, and then increases again as shown in training iteration 0 to about $0.25 \times 10^5$. This tendency is repeated until the agent converges.
As shown in Fig.~\ref{fig:reward128}, the performance of the agent also increases in all cases. As the sampled directions used in the training procedure increases, the performance increases quickly. However, the agent converges at higher rewards compared to the other cases. The agent trained with $N = 64$ has converged to $13258$. Note that the agent with $N = 64$ has higher reward than the agent trained with $N = 128$ until about $2\times10^5$ training iteration. This shows that the characteristic of the RAIL that optimizes the policy with the randomly generated noise. The generated noise sometimes finds the suboptimal policy quickly and converged to the suboptimal policy with a small number of sampled directions. 
When $N = 128$, the training curve increases and converges to $20453$. Note that the training curve increases continuously from 0 to approximately $1\times10^5$ training iteration when $N = 512$. The tendency for this performance to increase is longer than in other cases. This is because the policy does not fall into the suboptimal where other cases converge through the agent finds updates direction with a lot of noise. As a result of the training procedure, the agent has converged to $28569$.
Fig.~\ref{fig:rewardtot} shows the learning curves for all cases mentioned above. The relative performance of the agents which are converged is presented.
As shown in Fig.~\ref{fig:rewardtot}, the RAIL has the ability to imitate the autonomous driving control problem effectively; and thus the agent trained with the RAIL is capable of establishing optimal driving strategies.
 
\subsection{Weights Visualization}
\begin{figure*}[t!]
\begin{subfigure}{.24\textwidth}
  \centering
    \includegraphics[width=1\columnwidth]{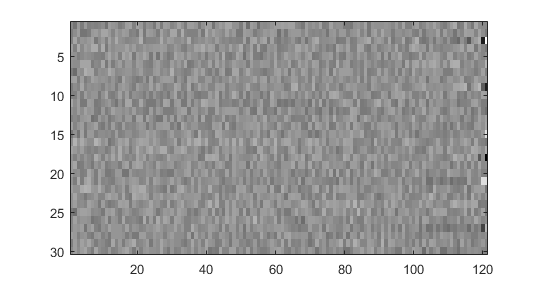}
    \caption{Weights from first layer trained through BC.}
    \label{fig:bcweight}
\end{subfigure}
\begin{subfigure}{.24\textwidth}
  \centering
    \includegraphics[width=1\columnwidth]{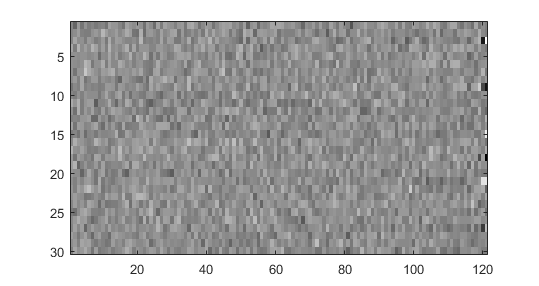}
    \caption{Weights from first layer trained through RAIL.}
    \label{fig:railweight}
\end{subfigure}
\begin{subfigure}{.24\textwidth}
  \centering
    \includegraphics[width=1\columnwidth]{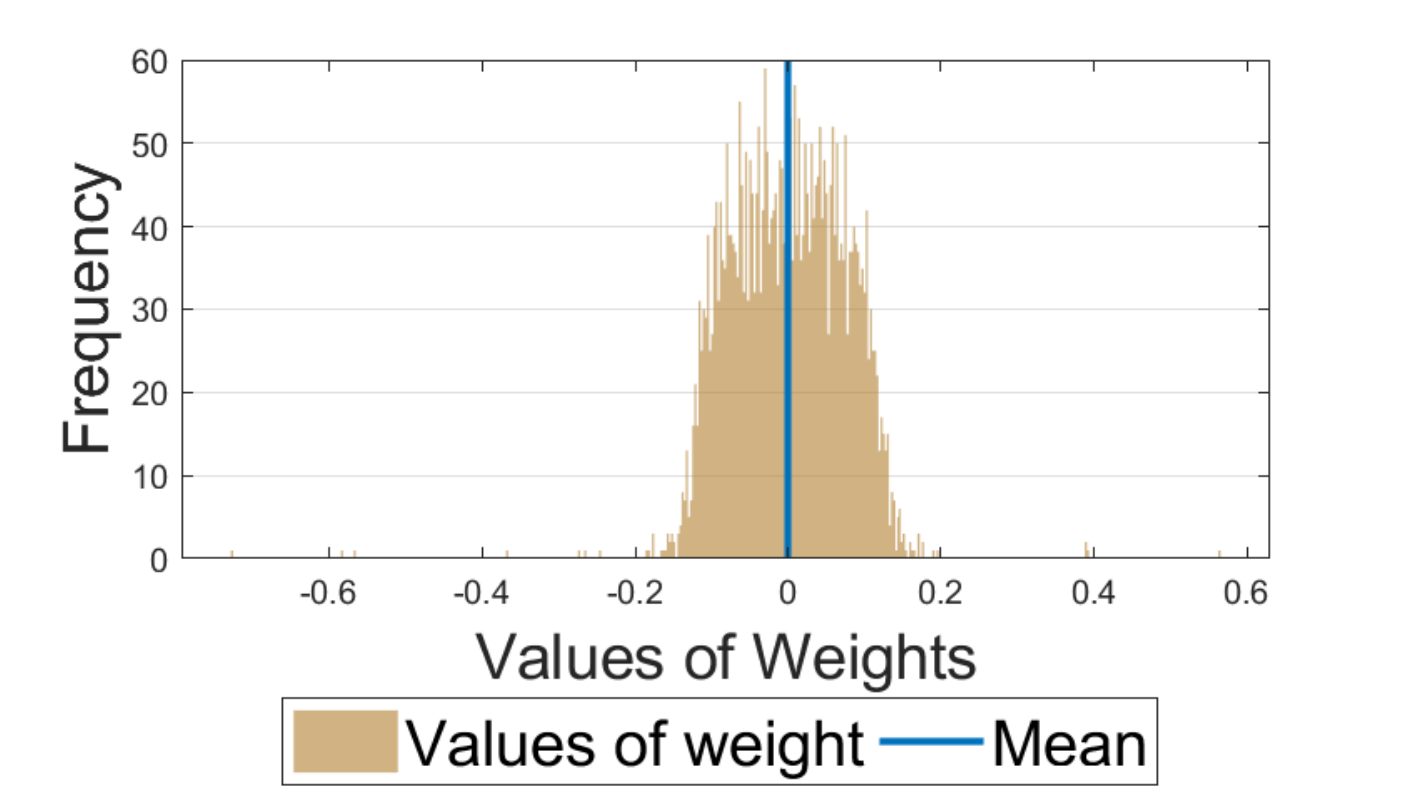}
    \caption{Weights distribution from first layer trained through BC.}
    \label{fig:bcweightdist}
\end{subfigure}
\begin{subfigure}{.24\textwidth}
  \centering
    \includegraphics[width=1\columnwidth]{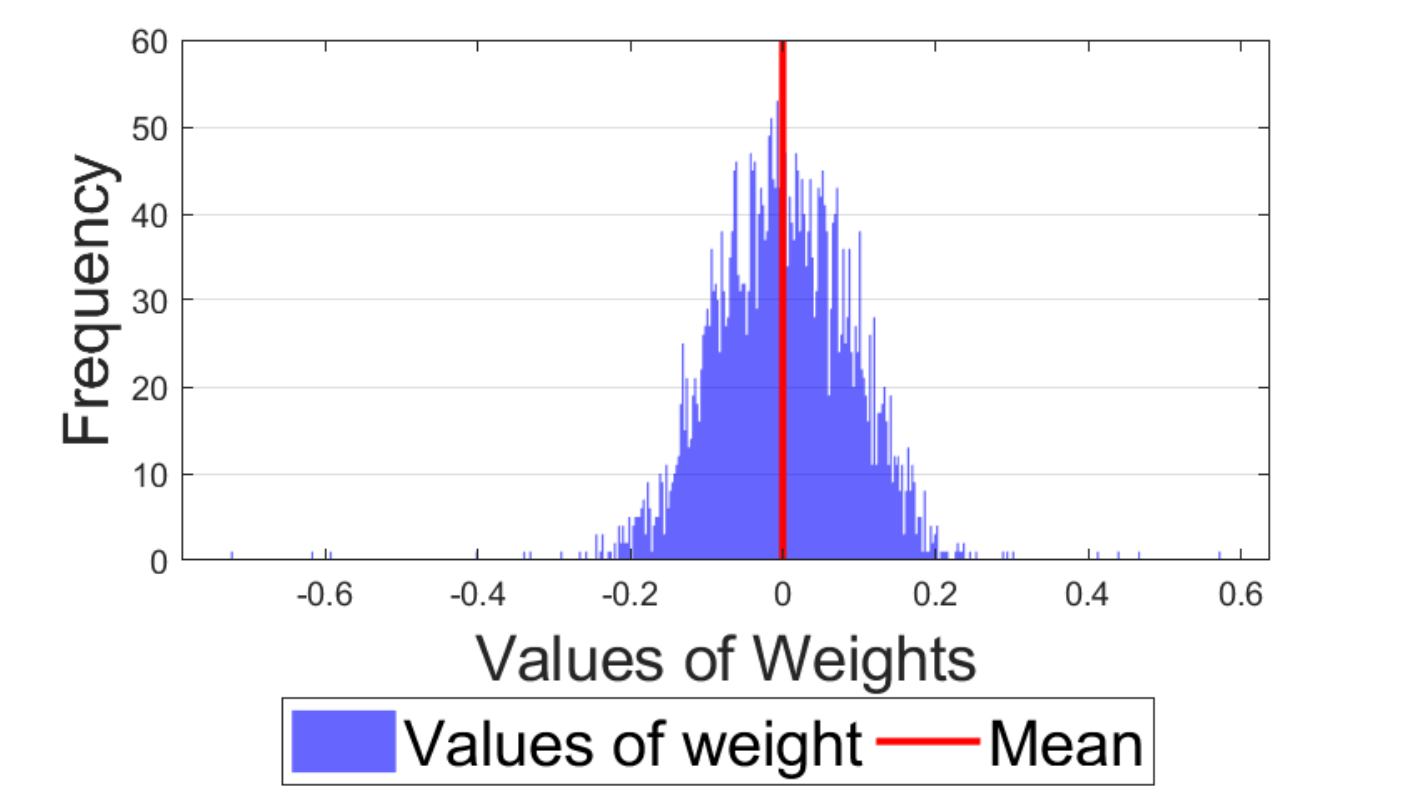}
    \caption{Weights distribution from first layer trained through RAIL.}
    \label{fig:railweightdist}
\end{subfigure}
\caption{The comparison of weights from first layer of the two-layer agent.}
\label{fig:weights}

\begin{subfigure}{.24\textwidth}
  \centering
    \includegraphics[width=1\columnwidth]{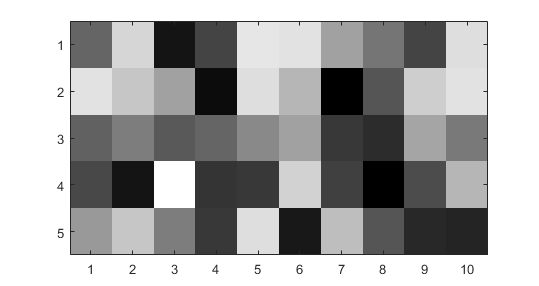}
    \caption{Weights from second layer trained through BC.}
    \label{fig:bcweight2}
\end{subfigure}
\begin{subfigure}{.24\textwidth}
  \centering
    \includegraphics[width=1\columnwidth]{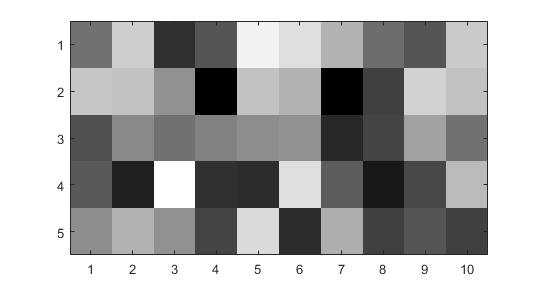}
    \caption{Weights from second layer trained through RAIL.}
    \label{fig:railweight2}
\end{subfigure}
\begin{subfigure}{.24\textwidth}
  \centering
    \includegraphics[width=1\columnwidth]{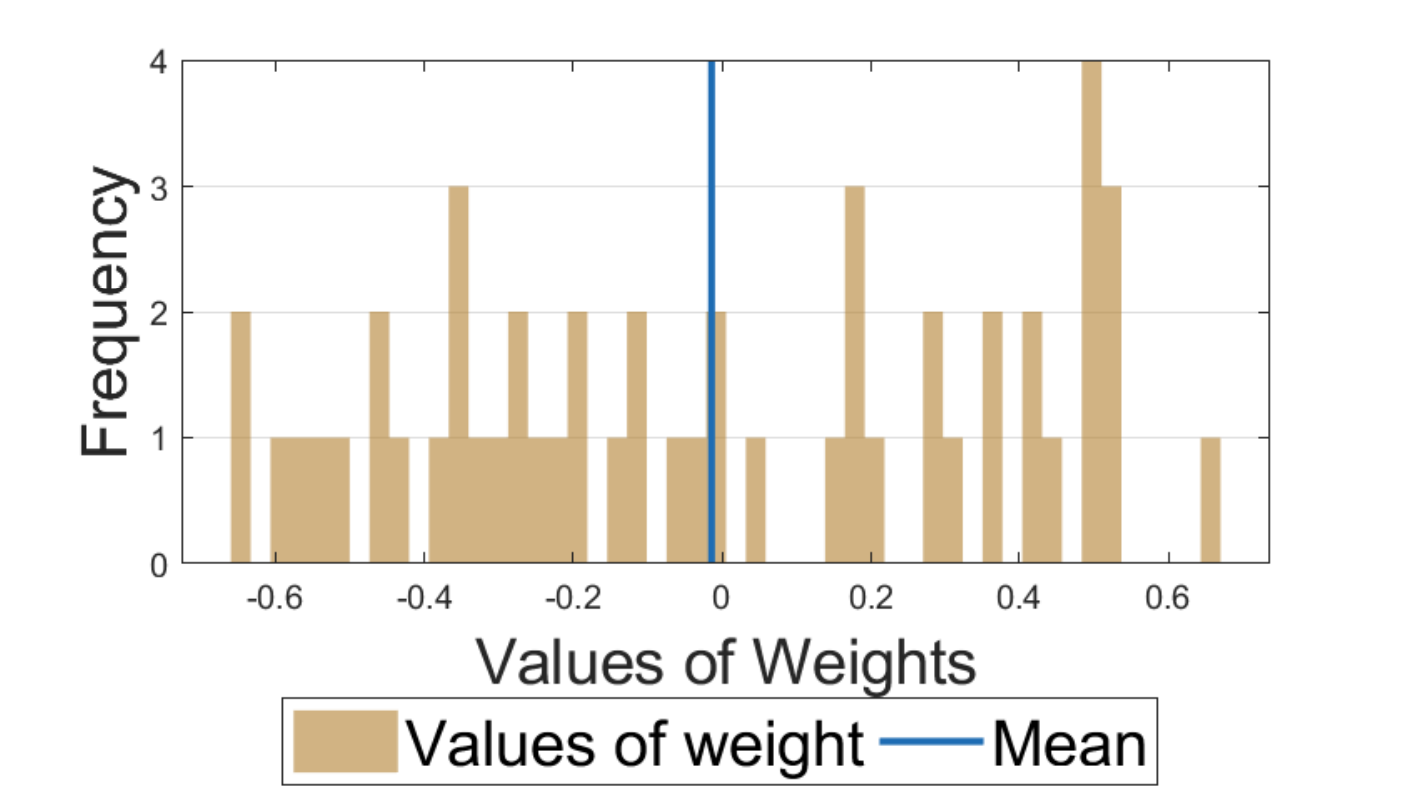}
    \caption{Weights distribution from second layer trained through BC.}
    \label{fig:bcweight2dist}
\end{subfigure}
\begin{subfigure}{.24\textwidth}
  \centering
    \includegraphics[width=1\columnwidth]{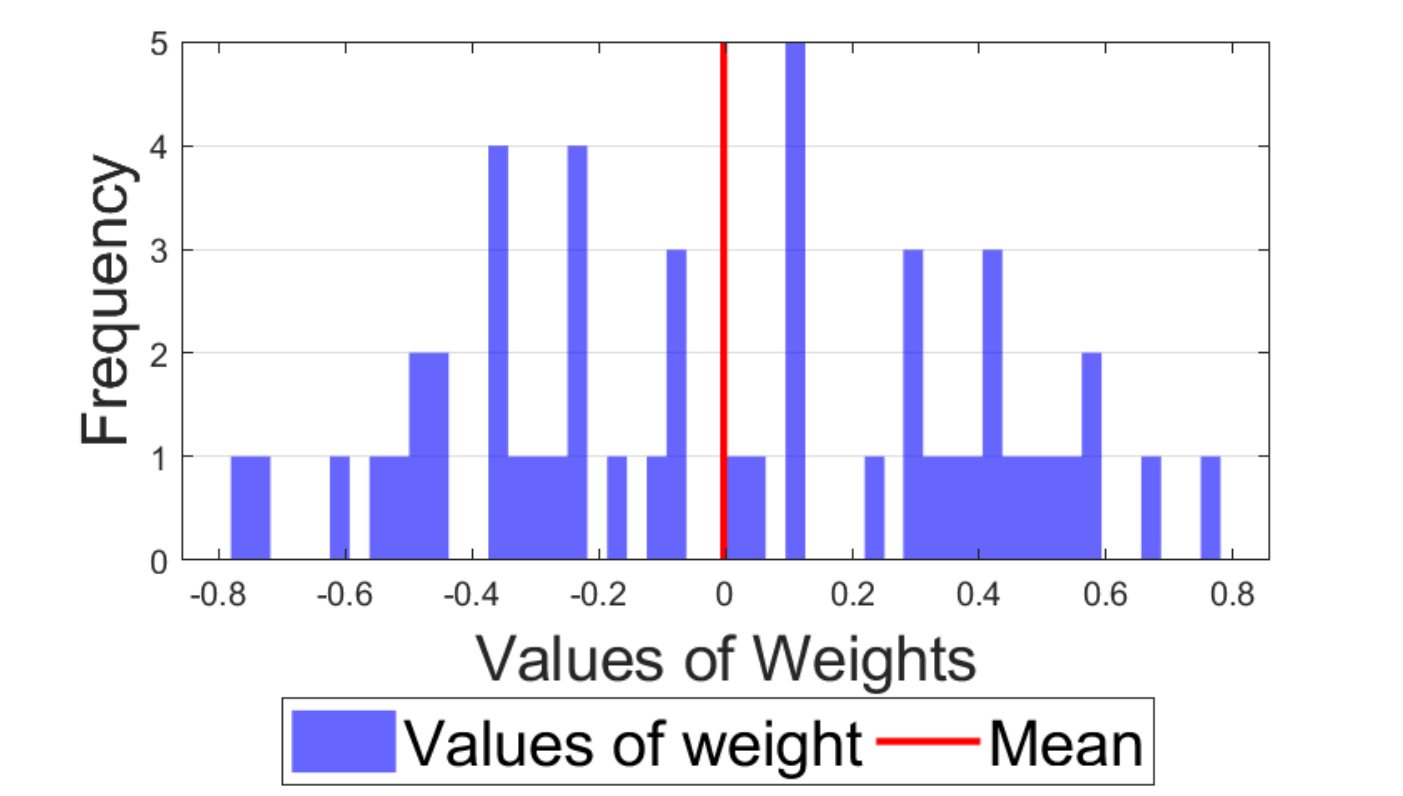}
    \caption{Weights distribution from second layer trained through RAIL.}
    \label{fig:railweight2dist}
\end{subfigure}
\caption{The comparison of weights from second layer of the two-layer agent.}
\label{fig:weights2}
\end{figure*}

This section compares the weights of the trained agents. As shown in Fig.~\ref{fig:sample}, the training methods influence the performance of the trained agent. The performance of the agent is closely related to the weights of the agent. Fig.~\ref{fig:weights} shows the trained weights for following two methods: behavior cloning (BC) and RAIL. As mentioned before, we use BC-based initialization to accelerate the training procedure. The weights of agents trained with RAIL is optimized by using the weights which are trained through BC. During the training procedure, the same settings are used in the subsection~\ref{sec:layer}.  Fig.~\ref{fig:weights} shows the weights in two-layer agent. The shape of weights from first layer is $10\times363$. We reshape the weights for visualization to $30\times121$ as shown in Fig.~\ref{fig:bcweight} and Fig.~\ref{fig:railweight}. The shape of weights from second layer is $5\times10$ as shown in Fig.~\ref{fig:bcweight2} and Fig.~\ref{fig:railweight2}. Therefore, as the value becomes low, the color becomes dark.

As shown in Fig.~\ref{fig:bcweight}, there are very high or low values compared to the other values of weights (i.e, 0.6 and -0.6). These values also exist in the trained weights through RAIL as shown in Fig.~\ref{fig:railweight} and Fig.~\ref{fig:railweightdist}. Similarly, in the weights in the second layer, very high or low values are approximately equivalent as shown in Fig.~\ref{fig:bcweight2} and Fig.~\ref{fig:railweight2}. This means that BC-based initialization to accelerate the training procedure is affected by these values. After the training procedure through RAIL, the weights have high or low values that do not exist as a result of BC. As mentioned before, BC trains a policy under the concept of supervised learning by observing the state-action pairs obtained by expert demonstration. Therefore, the differences between Fig.~\ref{fig:bcweight} and Fig.~\ref{fig:railweight} mean the results of RAIL that interacts with the environment to train the policy. These differences in weights make the different results in the above experiments as presented in Sec.~\ref{sec:layer}.

\section{Concluding Remarks}\label{sec:conclusion}
In this paper, a randomized adversarial imitation learning (RAIL) algorithm is proposed for effect autonomous driving policy training that coordinates ADAS functions for guaranteeing vehicles' safety. The proposed RAIL algorithm is derivative-free as well as model-free. With the proposed RAIL algorithm, the driving policies which efficiently and successfully control autonomous vehicle driving are trained via derivative-free optimization. While conducting the policy training, the simple randomized updates let the algorithm be facile; and therefore it makes the policy reconstruction computation results which achieve superior improved performance. By comparing the proposed RAIL algorithm with complex deep reinforcement learning inspired methods, it is observed that the proposed RAIL algorithm conducts policy training that achieves desired performance improvements. 
Furthermore, the experiments show that how BC-initialization accelerates the training procedure. 
This results can be one of the most promising ways to the common belief that randomized search based methods in the parameter spaces of autonomous driving policies can not be competitive. The performance evaluation results show the possibility that the ADAS functions of the autonomous vehicles can be coordinated and controlled by the policies derived by the proposed RAIL algorithm.

\bibliographystyle{IEEEtran}

\begin{IEEEbiography}[{\includegraphics[width=1in,height=1.25in,clip]{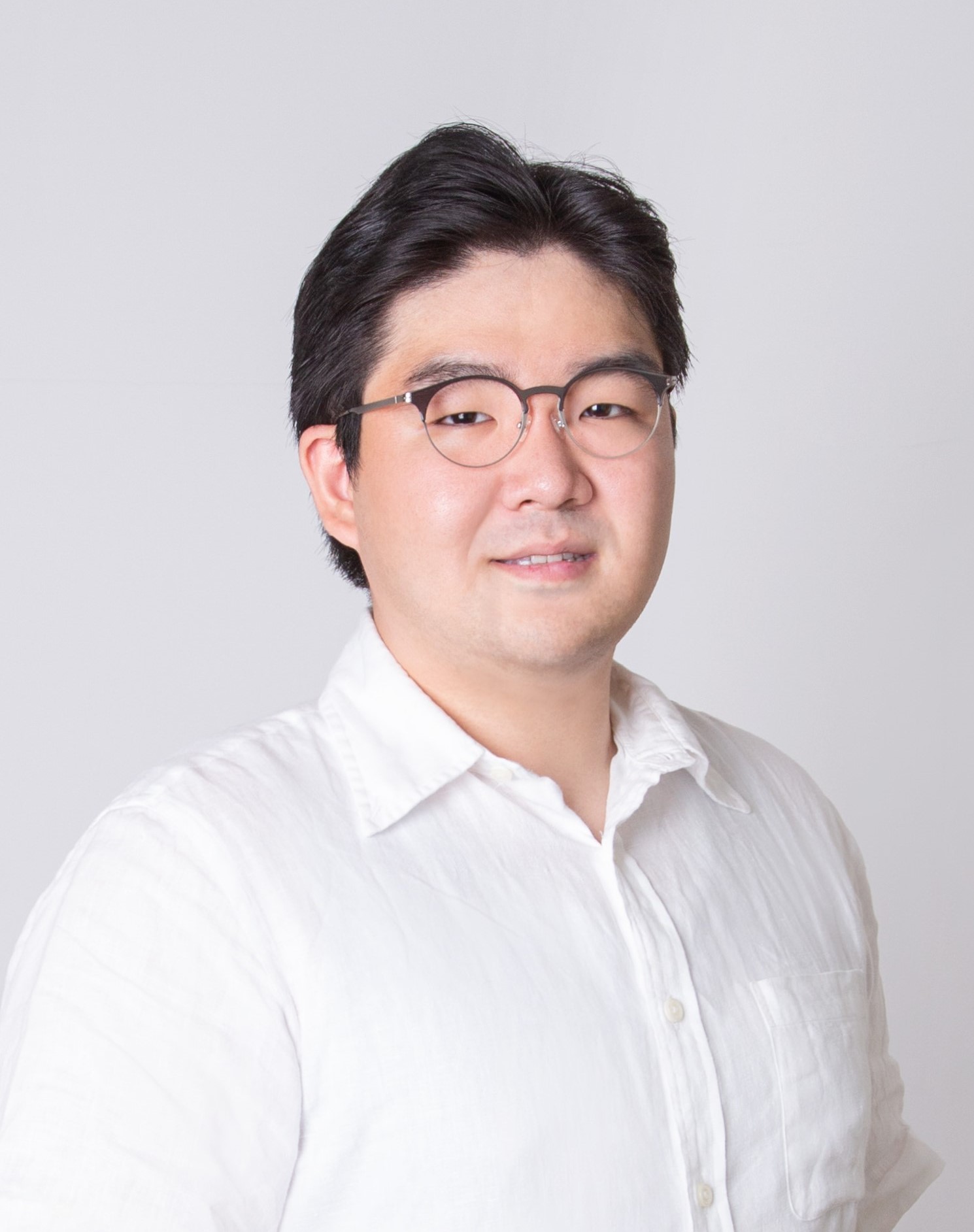}}]{Won Joon Yun}
is currently a Ph.D. student in electrical and computer engineering at Korea University, Seoul, Republic of Korea, since March 2021, where he received his B.S. in electrical engineering. His current research interests include multi-agent deep reinforcement learning for various mobile and network systems. His current research interests include multi-agent deep reinforcement learning for various mobile and network systems. He was a recipient of the Best Paper Awards by KICS (2020--2021) and IEEE ICOIN Best Paper Award (2021).
\end{IEEEbiography}
	
\begin{IEEEbiography}[{\includegraphics[width=1in,height=1.25in,clip]{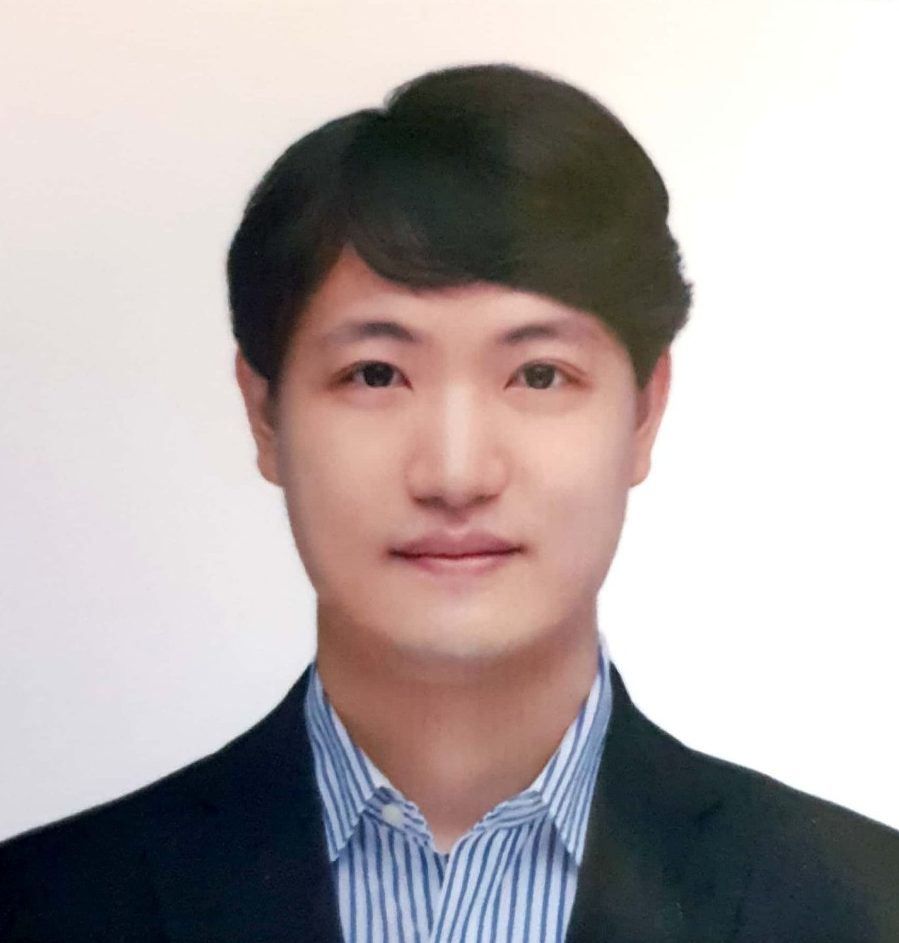}}]{MyungJae Shin} is currently an AI researcher at Mofl Inc., Daejeon, Republic of Korea.
He received his B.S. (\textit{second highest honor} from the College of Engineering) and M.S. degrees in computer science and engineering from Chung-Ang University, Seoul, Korea, in 2018 and 2020, respectively. 

His research interests are in various econometric theories and their deep-learning based computational solutions. He was a recipient of National Science \& Technology Scholarship (2016--2017) and IEEE Vehicular Technology Society (VTS) Seoul Chapter Award (2019).
\end{IEEEbiography}

\begin{IEEEbiography}[{\includegraphics[width=1in,height=1.25in,clip]{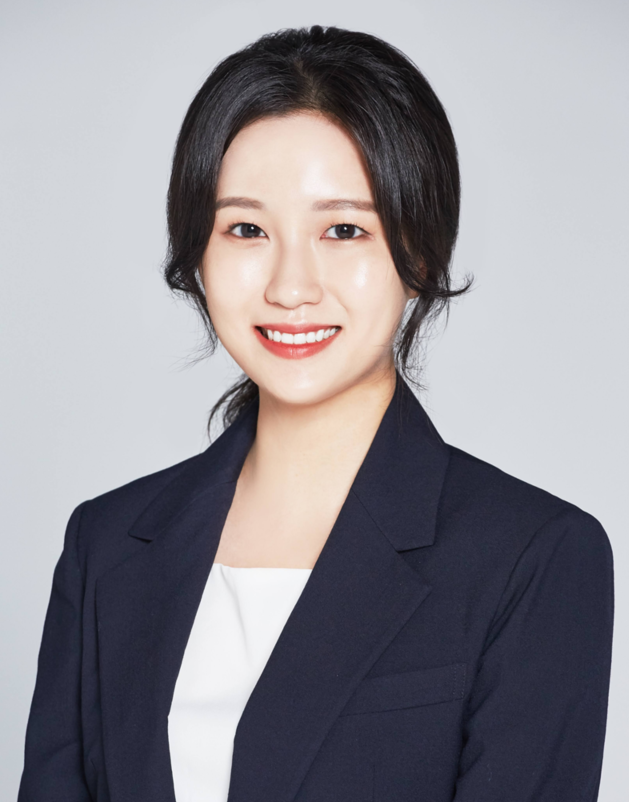}}]{Soyi Jung} has been an assistant professor at the School of Software, Hallym University, Chuncheon, Republic of Korea, since September 2021. She also holds a visiting scholar position at Donald Bren School of Information and Computer Sciences, University of California, Irvine, CA, USA, from 2021 to 2022. She was a research professor at Korea University, Seoul, Republic of Korea, during 2021. She was also a researcher at Korea Testing and Research (KTR) Institute, Gwacheon, Republic of Korea, from 2015 to 2016. 
She received her B.S., M.S., and Ph.D. degrees in electrical and computer engineering from Ajou University, Suwon, Republic of Korea, in 2013, 2015, and 2021, respectively. 

Her current research interests include network optimization for autonomous vehicles communications, distributed system analysis, big-data processing platforms, and probabilistic access analysis. She was a recipient of Best Paper Award by KICS (2015), Young Women Researcher Award by WISET and KICS (2015), Bronze Paper Award from IEEE Seoul Section Student Paper Contest (2018), ICT Paper Contest Award by Electronic Times (2019), and IEEE ICOIN Best Paper Award (2021).
\end{IEEEbiography}

\begin{IEEEbiography}[{\includegraphics[width=1in,height=1.25in,clip]{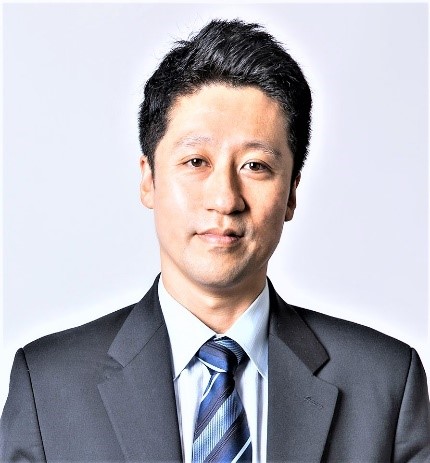}}]{Sean Kwon} (S'09-M'14) received his Ph.D. degree from Georgia Institute of Technology in Atlanta, US on December 2013. Before that, Dr. Kwon received the M.Sc degree from the University of Southern California, Los Angeles, US in 2007; the B.Sc degree from Yonsei University, Seoul, South Korea in 2001.
He performed research at Intel's Next Generation and Standards Division in Communication and Devices Group during 2015 to 2017, where he contributed to 5G MIMO standards, the associated system design and patents. He has been Assistant Professor at California State University Long Beach; and Chief Director/Founder of Wireless Systems Evolution Laboratory (WiSE Lab) since 2017.

He also conducted postdoctoral research at Wireless Devices and Systems Group, University of Southern California in 2014 - 2015. He worked on CDMA common air interface focusing on layer-3 protocols at the R$\&$D Institute of Pantech co., Ltd, Seoul, South Korea in 2001 to 2004. He was involved in several projects such as a DARPA project; an US Army Research Lab project; and 6 mobile-station projects for Motorola and Sprint, which were successfully on the market. His current research interests are in 5G and beyond-5G wireless system/network design; satellite communications; polarization diversity and multiplexing; body area network such as wearable computing; wireless channel modeling and its applications; and application of machine learning for wireless communications and signal processing.

He was a recipient of 3 Best Paper Awards from IEEE Green Energy and Smart Systems Conference (IGESSC), 2018, 2019 and 2020.
\end{IEEEbiography}
	
\begin{IEEEbiography}[{\includegraphics[width=1in,height=1.25in,clip]{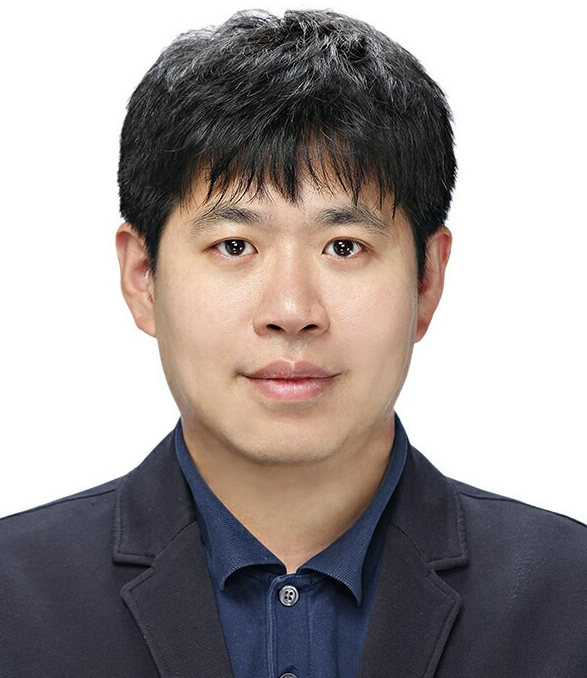}}]{Joongheon Kim}
(M'06--SM'18) has been with Korea University, Seoul, Korea, since September 2019, and he is currently an associate professor at the School of Electrical Engineering. He is also a vice director of the Artificial Intelligence Engineering Research Center at Korea University, Seoul, Korea. 
He received the B.S. and M.S. degrees in computer science and engineering from Korea University, Seoul, Korea, in 2004 and 2006, respectively; and the Ph.D. degree in computer science from the University of Southern California (USC), Los Angeles, California, USA, in 2014. 
Before joining Korea University, he was with LG Electronics CTO Office, Seoul, Korea, from 2006 to 2009; InterDigital, San Diego, California, USA, in 2012; Intel Corporation, Santa Clara in Silicon Valley, California, USA, from 2013 to 2016; and Chung-Ang University, Seoul, Korea, from 2016 to 2019. 

He is a senior member of the IEEE and serves as an associate/guest editor for \textit{IEEE Transactions on Vehicular Technology}, \textit{IEEE Communications Standards Magazine}, and \textit{Computer Networks (Elsevier)}. He is also a distinguished lecturer for \textit{IEEE Communications Society} (2022--2023).

He was a recipient of the Annenberg Graduate Fellowship with his Ph.D. admission from USC (2009), 
Intel Corporation Next Generation and Standards (NGS) Division Recognition Award (2015), Haedong Young Scholar Award by KICS (2018), Paper Awards from IEEE Seoul Section Student Paper Contests (2019 and 2020), \textit{IEEE Systems Journal} Best Paper Award (2020), IEEE ICOIN Best Paper Award (2021), Haedong Paper Award by KICS (2021), and IEEE Vehicular Technology Society (VTS) Seoul Chapter Awards (2019 and 2021).
\end{IEEEbiography}

\end{document}